\pdfoutput=1

\documentclass[11pt]{article}

\usepackage[]{acl}

\usepackage{times}
\usepackage{latexsym}

\usepackage{booktabs}
\usepackage{graphicx}
\usepackage{amsmath, amssymb}

\usepackage[T1]{fontenc}

\usepackage[utf8]{inputenc}

\usepackage{microtype}

\usepackage{inconsolata}

\usepackage{multirow}
\usepackage{subcaption}

\newcommand{\igl}[1]{
    \textcolor{purple}{[IL: #1]}
}

\DeclareMathOperator*{\argmin}{arg\,min}

\title{On Retrieval Augmentation \\ and the Limitations of Language Model Training}

\author{Ting-Rui Chiang \quad Xinyan Velocity Yu \quad Joshua Robinson \\ \textbf{Ollie Liu} \quad \textbf{Isabelle Lee} \quad \textbf{Dani Yogatama} \\
  University of Southern California \\
  \texttt{\{tingruic,xinyany,joshua.j.robinson,zliu2898,gunheele,yogatama\}@usc.edu}
}

\begin{document}
\maketitle
\begin{abstract}

Augmenting a language model (LM) with $k$-nearest neighbors ($k$NN) retrieval on its training data alone can decrease its perplexity, though the underlying reasons for this remain elusive. In this work, we rule out one previously posited possibility --- the ``softmax bottleneck.'' We then create a new dataset to evaluate LM generalization ability in the setting where training data contains additional information that is not causally relevant. This task is challenging even for GPT-3.5 Turbo. We show that, for both GPT-2 and Mistral 7B, $k$NN retrieval augmentation consistently improves performance in this setting. Finally, to make $k$NN retrieval more accessible, we propose using a multi-layer perceptron model that maps datastore keys to values as a drop-in replacement for traditional retrieval. This reduces storage costs by over 25x.\footnote{The source code is available at \url{https://github.com/usc-tamagotchi/on-knnlm}.}

\end{abstract}

\section{Introduction}

Recent efforts to improve the performance of language models (LMs) have focused on scaling up model \citep{gpt3paper} and training data size \citep{Hoffmann2022TrainingCL}. The resulting models have reached near-human or even super-human performance on some tasks ~\citep{chowdhery2022palm}, though with steep accompanying energy and compute resource costs \citep{schwartz2020green, gpt3paper, touvron2023llama2}.

Another approach for improving LM performance has been retrieval augmentation. \citet{Khandelwal2020Generalization} 
proposed to build a datastore using LM training data.
The datastore associates the next token of prefixes in the training data with the representations of the prefixes extracted from an intermediate layer of an LM.
They found that when predicting the next token for a given prefix, using \textit{k}-nearest neighbor ($k$NN) retrieval, which retrieves the next token based on the intermediate representation of a given prefix, reduced language models' perplexity.
Because the datastore is drawn entirely from the LM's training data,
the success of $k$NN augmentation suggests the standard LM training setup does not yield models that best utilize their parametric capacity or training data. 
Studying why LMs augmented with $k$NN retrieval ($k$NN-LMs) outperform vanilla LMs may shed light on ways to improve the standard LM training setup.

In this work, we base our study on the analyses of $k$NN-LMs by \citet{why-knn}.
Among the aspects they explore are the limitations of model architecture and memorization. 
\citet{why-knn} suggest the $k$NN component may be able to map intermediate representation of context to distributions in a more flexible way, while the last layer of LMs has a softmax bottleneck~\citep{yang2018breaking} that restricts LMs from generating certain distributions. 
This discrepancy of expressiveness may thus cause the performance gap. 
They also show that replacing the $k$NN component with an overfitted LM performs worse than $k$NN-LM, suggesting that $k$NN augmentation does not perform better solely because it memorizes the training data better.

In this work, we start with inspecting the bottlenecks in the model as suggested by \citet{why-knn}.
We propose an experiment that shows that the softmax bottleneck is not the cause of the performance gap between $k$NN and vanilla LM.
Our experimental results show that the last linear layers of LMs can generate distributions that approximate the distribution from a $k$NN-LM well.
Therefore, we conclude that the bottleneck issues in the last layers, including the softmax bottleneck issue, are not the cause of the performance gap. 

We then investigate the performance gap from the perspective of generalization.
This explains why an overfitted LM is less effective than a $k$NN retrieval component~\citep{why-knn}.
We identify a scenario which we refer to as \textit{over-specification}.
That is, when a statement about certain knowledge (e.g., relational knowledge~\citep{petroni-etal-2019-language} or commonsense~\citep{conceptnet,aaai-commonsense,atomic}) contains redundant information.
We create a synthetic dataset \textit{Macondo} and use it to show that over-specification in training data prevents LMs from learning the knowledge in a robust way, i.e., LMs cannot generalize to test data which is not over-specified.
Even GPT-3.5 Turbo, fails, indicating it is a fundamental limitation of LM training.
It may be crucial when the size of training data is limited, because in this scenario, it is likely that there are only few statements about certain knowledge and all of them are over-specified. Decounfounding the effect of having redundant information also requires more training examples. 
This may explain why we need to scale up the training data size.

Because the better generalization ability may be what makes the $k$NN component helpful, we explore alternatives to a $k$NN component by looking for components that also generalize well.
It turns out that we can close the generalization gap on Moncodo by training another neural model that maps the intermediate representation to the target token.
We also show that on the WikiText dataset, this approach reduces the perplexity by 1.45 while requiring less than 4\% storage space of $k$NN augmentation.
We suggest it is a promising future direction for improving LMs.

\section{Background and Notations}
\label{sec:background}


\paragraph{LM} We focus on Transformer LMs such as GPT-2. Given context $c = \{ x_i \}_{i=1}^{t-1}$, we formulate next token prediction as
\begin{equation}
    p_{\text{lm}}(x_t | c ) = f \circ \mathrm{g} \circ \mathrm{enc}(c),
    \label{eq:lm}
\end{equation}
where $f$ is the last linear layer with softmax activation, $g$ is the two-layer MLP network with a residual connection in the last Transformer layer, and $\mathrm{enc}$ includes the earlier layers of the model. 

\paragraph{$k$NN-LM} \citet{Khandelwal2020Generalization} use the $\mathrm{enc}$ function from a trained LM (Eq~\ref{eq:lm}) to build a datastore, where a key is the representation of a token sequence $\{ x_i \}_{i=1}^{t-1}$ in the training data encoded by $\mathrm{enc}$, and the value of the key is the next token $x_t$. 
When predicting the next token $x'_t$ of given context $c = \{ x'_i \}_{i=1}^{t-1}$, $k$NN-LM has
a $k$NN retrieval module that maps $\mathrm{enc}(c)$ to a distribution $p_{\text{knn}}(\cdot | c)$ by querying the datastore with $\mathrm{enc}(c)$.
Then a $k$NN-LM generates the next token distribution with 
\begin{equation*}
    p_{\text{knnlm}}(x_t | c) = \lambda p_{\text{lm}}(x_t | c ) + (1 - \lambda) p_{\text{knn}}(x_t | c),
\end{equation*}
where $\lambda$ is a hyperparameter for interpolation.


\paragraph{Softmax bottleneck}
\citet{yang2018breaking} theoretically show that the dimensionality of the last linear layer confines the possible vocabulary distribution the last softmax layer can generate.
It implies that no matter what $g \circ \mathrm{enc}$ generates, $f$ can not generate certain distributions.

\section{Capacity of LMs' Last Layers}
\label{sec:bottleneck}

\citet{why-knn} hypothesize that the performance gap between $k$NN-LM and vanilla LM is because the softmax bottleneck prevents it from generating some distributions that $k$NN-LM can generate.
In this section, we reinspect this hypothesis. 

\subsection{Projecting to the Probability Space} 
\label{subsec:projection}
We study whether softmax bottleneck causes the performance gap by inspecting whether the last layers can generate a distribution that approximates the distribution generated by $k$NN-LM $p_{\text{knnlm}}$.
We do the projection by solving
\begin{equation}
    z^* \in \argmin_{z \in \mathbb{R}^d} \mathrm{KL}[ f(z) \| p_{\text{knnlm}} ],
    \label{eq:project}
\end{equation}
where $f$ is the last layer of the model with its trained parameters fixed (definition in Eq~\ref{eq:lm}).
By definition, if softmax bottleneck really prevents the model from generating $p_{\text{knnlm}}$, then $f(z^*)$ can not approximate $p_{\text{knnlm}}$ well and thus its perplexity should be close to the vanilla LM's.
Therefore, by comparing the perplexity of $p_{\text{proj}} = f(z^*)$ with vanilla LM's and $k$NN-LM's perplexity, we can infer the effect of softmax bottleneck in this problem.

Similarly, we can inspect whether the MLP layer has a bottleneck effect by replacing $f$ in Eq~\ref{eq:project} with $f \circ g$.
We use $\mathrm{enc}(\{ x_i \}_{i=1}^{t-1})$ as the initialization of $z$ and solve Eq~\ref{eq:project} with gradient descent.

\begin{table}[]
    \centering
    \begin{tabular}{cc|cc}
        \toprule
        \multicolumn{2}{c|}{Original LMs} & \multicolumn{2}{c}{$p_{\mathrm{proj}}$ projected with Eq.~\ref{eq:project}} \\
        LM & $k$NN-LM & $f \to p_{\text{knnlm}}$ & $f \circ g \to p_{\text{knnlm}}$ \\
        \midrule
         20.13  & 16.92  &  16.76 & 16.78 \\
        \bottomrule
    \end{tabular}
    \caption{The perplexity of the LMs discussed in \S{\ref{sec:bottleneck}}.
    }
    \label{tab:perp}
\end{table}

\subsection{Experiment, Result, and Discussion}
We train an LM using WikiText following the setting in~\citet{Khandelwal2020Generalization} and measure its perplexity (details in \S\ref{sec:bottleneck-details}).
Table~\ref{tab:perp} shows that the approximation of $p_{\text{knnlm}}$ by the last layer $f$ has an average perplexity similar to the perplexity of $k$NN-LM.
The average KL-divergence between $p_{\mathrm{knnlm}}$ and $p_{\mathrm{proj}}$ is also under 0.1 (Table~\ref{tab:kld}). 
These results imply that the approximation is good enough for a good perplexity.
It also implies the softmax bottleneck does not prevent the LM from generating a good distribution.
Thus, the softmax bottleneck is not the cause of the gap between vanilla LM and $k$NN-LM.
Projecting the $p_{\mathrm{knnlm}}$ to the output space of $f \circ g$ has a similar result.
Therefore, LMs' last layers do not have a bottleneck that causes the performance gap.
\footnote{However, we find it more difficult to solve Eq.~\ref{eq:project} with a smaller learning rate for $f \circ g$. More discussions in \S\ref{sec:mlp-hurdle}.}

\section{Generalization from Over-specification}
\label{sec:over-spec}

As the last layers do not have a bottleneck issue that explains the performance gap, we turn to study the efficacy of $k$NN augmentation from the perspective of generalization.
In this section, we identify a limitation of LM training that may cause the performance gap: The failure to generalize from \textit{over-specified} descriptions.

\subsection{Over-specification}

We refer to the phenomenon that the prefix of a partial sentence contains information that is not causally relevant to its completion as \textit{over-specification}.
In other words, over-specification is the scenario where removing some information in the prefix (e.g. a phrase) does not change the likelihood of the continuation.
This phenomenon often occurs in in the training data. 
The descriptions about factual knowledge or commonsense are usually over-specified with non-causally relevant information, but the causally irrelevant information may be absent during inference.
Generalization from over-specified training data is thus important for an LM to utilize knowledge in the training data.

For example, in the training data, the text about the knowledge ``being drunk'' implies ``dangerous to drive'' may be over-specified as ``I was drunk when I left the party, so it was dangerous to drive''. 
In this example, ``I was drunk'' is causally relevant to ``it was dangerous to drive'' but ``when I left the party'' is not. An ideal LM should generalize and predict the same continuation when the non-causal information ``when I left the party'' is absent.

\subsection{Dataset: Macondo }
We create a synthetic dataset Macondo to demonstrate the challenge of generalizing from over-specified training data.
This dataset contains the names of 500 villagers, where 100 villagers have 1 to 5 child(ren), and each villager has a unique full name consisting of a random combination of a first name and a last name. Each child has a single-token and distinct first name. 
We construct each sentence in the training set using the template ``\texttt{[villager]}, who \texttt{[desc]}, is the parent of \texttt{[child]}'', where ``\texttt{[desc]}'' is a verb phrase about an attribute of the villager that is irrelevant to the parent-child relationship.
As for the sentences in the test set, they follow the template ``\texttt{[villager]}, is the parent of \texttt{[child]}''.
A perfect LM should predict each child of the villager with probability $\log(1/ \text{\# of children})$.
(More details in \S\ref{sec:macondo-ds})

\subsection{Experiment, Results, and Discussion}
\label{sec:macondo-result}

\begin{figure}[h]
    \centering
    \includegraphics[width=\linewidth]{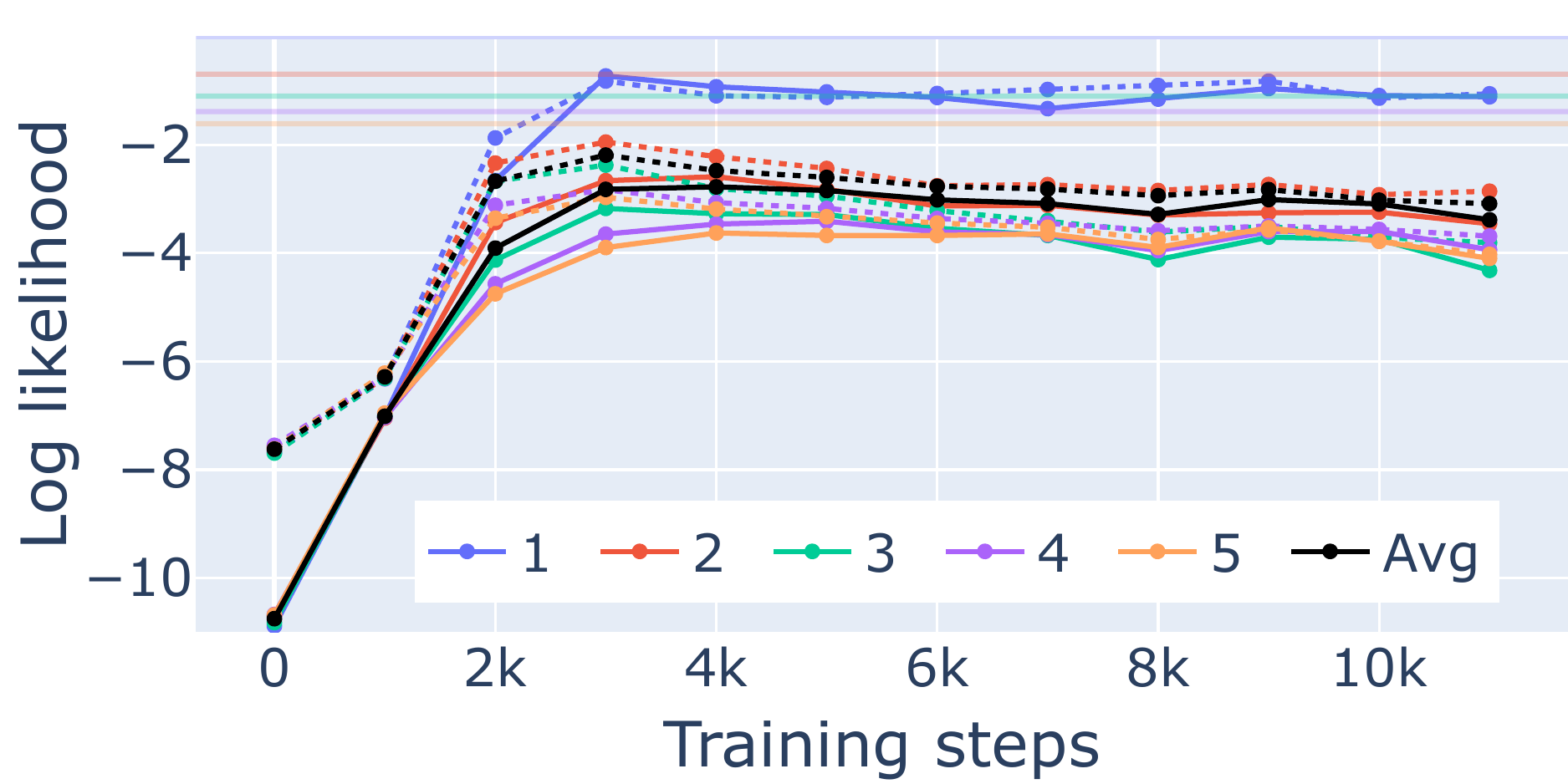}
    \caption{Test log-likelihood of children names in our synthetic dataset Macondo predicted by a fine-tuned GPT-2 XL model for parents with 1-5 children (average of \textbf{5} random seeds).
    The dotted lines represent the results of the $k$NN augmented LM. 
    The horizontal lines represent the theoretically best log-likelihood a perfect model can achieve ($\log(1/ \text{\# of children})$). 
    See Table~\ref{table:macondo-gpt2-xl} for the exact statistics shown in this figure.}
    \label{fig:macondo-gpt2-xl}
\end{figure}

To inspect how LMs are (un)able to generalize from over-specified training data, we fine-tune GPT-2 XL models with Macondo and test it with the test set where irrelevant ``\texttt{[desc]}'' is absent (details in \S\ref{sec:macondo-exp-detail}).
Figure~\ref{fig:macondo-gpt2-xl} shows that the fine-tuned GPT-2 model has a likelihood much lower than the theoretical perfect likelihood ($\log(1/ \text{\# of children})$).
It indicates that it cannot generalize from over-specification.
Additionally, Figure~\ref{fig:macondo-gpt2-xl} shows that the $k$NN-augmented model performs better than the vanilla model.
The better generalization capability of an augmented LM may partly explain the performance gap between augmented and vanilla LMs.
We also experiment with GPT-2 small models (Figure~\ref{fig:macondo-gpt2}) and find that GPT-2 XL models do not generalize much better, suggesting that scaling up the model may not close this generalization gap. We observe similar performance trend when fine-tuning a Mistral-7B-v0.1 model in Section \ref{sec:mistral-result}.

\subsection{Experimenting with GPT-3.5-turbo}
To inspect whether scaling mitigates the challenge of generalization, we further experiment with GPT-3.5-turbo.
We construct a conversational version of Macondo, Macondo-Conv to fit the conversational format of GPT-3.5-turbo.
In the training set, sentences follow the template ``User: Who is the child of \texttt{[villager]}, the one who \texttt{[desc]}? Assistant: \texttt{[child]}.''.
The test examples follow the template ``User: Who is the child of \texttt{[villager]}? Assistant: \texttt{[child]}.''.
The dataset contains 125 villagers having 2 children for lower fine-tuning costs. 

\begin{figure}
    \centering
    \includegraphics[width=\linewidth]{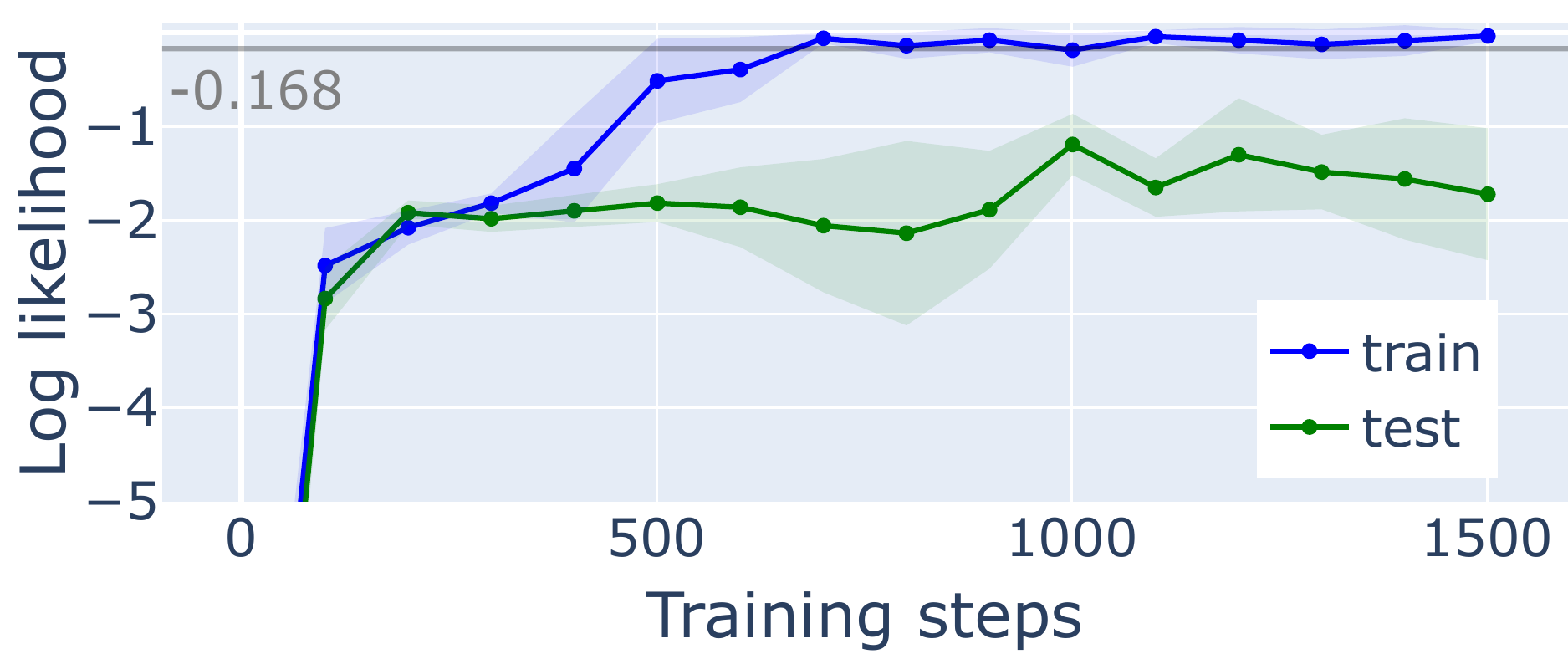}
    \caption{GPT-3.5-turbo fine-tuned with Macondo-Conv using OpenAI API. 
    The results are the average of 5 runs with 5 datasets generated with 5 random seeds. 
    Note that the presented loss involves special tokens, e.g., end-of-string tokens, so the theoretical perfect likelihood is greater than $\log 0.5$. The gray line is the test loss we achieve when we use the test data to train the model.}
    \label{fig:gpt35}
\end{figure}

The result in Figure~\ref{fig:gpt35} shows that GPT-3.5-turbo can not generalize to a test set without over-specification. 
This suggests that scaling up the model size alone cannot solve this generalization challenge.
This failure to generalize may be a fundamental limitation of LM training.

\begin{table}[]
    \centering
    \begin{tabular}{cc|cc}
        \toprule
        \multicolumn{2}{c|}{Macondo} & \multicolumn{2}{c}{WikiText} \\
        LM & $k$NN/MLP-LM & LM & $k$NN/MLP-LM   \\
        \midrule
         19.66  & 17.69 / 10.76  &  20.13 & 16.92 / 18.68 \\
        \bottomrule
    \end{tabular}
    \caption{The perplexity of LMs augmented with different a $k$NN model or a MLP model (\S\ref{sec:mlp-aug}).
    }
    \label{tab:mlp-aug}
\end{table}

\section{An Alternative to $k$NN-augmentation}
\label{sec:mlp-aug}

Motivated by the results in \S\ref{sec:macondo-result}, we explore whether using a datastore is necessary to improve perplexity. 
The success of $k$NN-augmentation in \S\ref{sec:macondo-result} shows that it is possible to generalize better by utilizing the intermediate representation with a $k$NN module.
We wonder whether we can use a classification model instead of a $k$NN module. 

Because deep models have been known for their generalization ability \citep{DBLP:journals/corr/NeyshaburTS14,neyshabur2018the},
we explore using an MLP model to replace $k$NN retrieval.
We use the key-value pairs in the datastore for $k$NN retrieval to train an MLP model to map the keys to the values (details in \S\ref{sec:mlp-aug-detail}).
Results in Table~\ref{tab:mlp-aug} show that interpolating the original LM with this MLP model effectively reduces the perplexity while requiring less than 4\% of storage.
This indicates a promising future direction.

\section{Related Work}
LMs that solely rely on parametric knowledge learned during training time are known to hallucinate~\citep{shuster-etal-2021-retrieval-augmentation, dhuliawala2023chain, zhang2023language, ye2023cognitive, zhang2023siren}, suffer to learn long-tail knowledge~\citep{roberts-etal-2020-much}, and fail to adapt to new knowledge over time~\cite{de-cao-etal-2021-editing, chen2021a, kasai2022realtime}. To overcome these limitations, recent works~\citep{Khandelwal2020Generalization, lewis2020retrieval, pmlr-v119-guu20a, yogatama-etal-2021-adaptive, borgeaud2022improving, JMLR:v24:23-0037, zhong-etal-2022-training, min-etal-2023-nonparametric} include an external datastore with the parametric model, resulting in a retrieval-augmentated model paradigm.  
Meanwhile, \citet{drozdov-etal-2022-cant} and \citet{wang2023knn} analyzes the effect of $k$NN-LM on generation tasks, while
\citet{shi-etal-2022-nearest} focuses on using $k$NN-LM on few- and zero-shot classification tasks.   

The traditional LM training setup has been shown to yield models that fail to generalize to test data with reversed relations \cite{berglund2023reversal}, respective readings \cite{cui-etal-2023-failure}, and longer tasks \cite{anil2022exploring}. These models can also struggle with linguistic generalization between unseen but related contexts \cite{10.1162/tacl_a_00608} and learn shortcuts that harm generalization \cite{mccoy-etal-2019-right}. \citet{bender-koller-2020-climbing} have also argued that such models will necessarily be limited due to the ungrounded nature of their training data.

\section{Conclusion}

We study the performance gap between vanilla and $k$NN-augmented LMs.
We develop an experiment that allows us to directly inspect the bottleneck issue and exclude the possibility that it causes the performance gap (\S{\ref{sec:bottleneck}}).
We further identify the over-specified scenario where vanilla LMs fail to generalize while $k$NN-LMs generalize better (\S{\ref{sec:over-spec}}).
We also show with GPT-3.5-turbo that this failure of generalization can not be solved by scaling up the model size, suggesting that this is a fundamental limitation of LM training.  
Finally, we show the potential of augmenting LMs with an MLP model, indicating a promising future direction (\S{\ref{sec:mlp-aug}}).

\section*{Limitations}

While we gain more insights by closely inspecting the phenomena observed by \citet{why-knn}, why $k$NN augmentation is beneficial remains not fully clear.
In \S\ref{sec:bottleneck}, we focus on the bottleneck issues of the last layers $f \circ g$ and show that there exists an intermediate representation $z^*$ such that $f \circ g(z^*)$ approximates $p_{\text{knnlm}}$ well. 
However, it is unclear why $\mathrm{enc}$ does not map the context to $z^*$.
In \S\ref{sec:over-spec}, we identify the over-specification scenario where $k$NN-LMs generalize better than vanilla LMs.
However, the mechanism behind this remains unclear.
In \S\ref{sec:mlp-aug}, we show that augmenting LMs with another MLP can improve the perplexity of the model but does not fully close the gap between $k$NN-LM and vanilla LM on WikiText.
Further analysis is required to understand the generalization behavior of the $k$NN and the MLP models.

\bibliography{anthology,custom}

\appendix

\begin{figure}[b]
    \centering
    \begin{subfigure}[b]{\linewidth}
        \includegraphics[width=\linewidth]{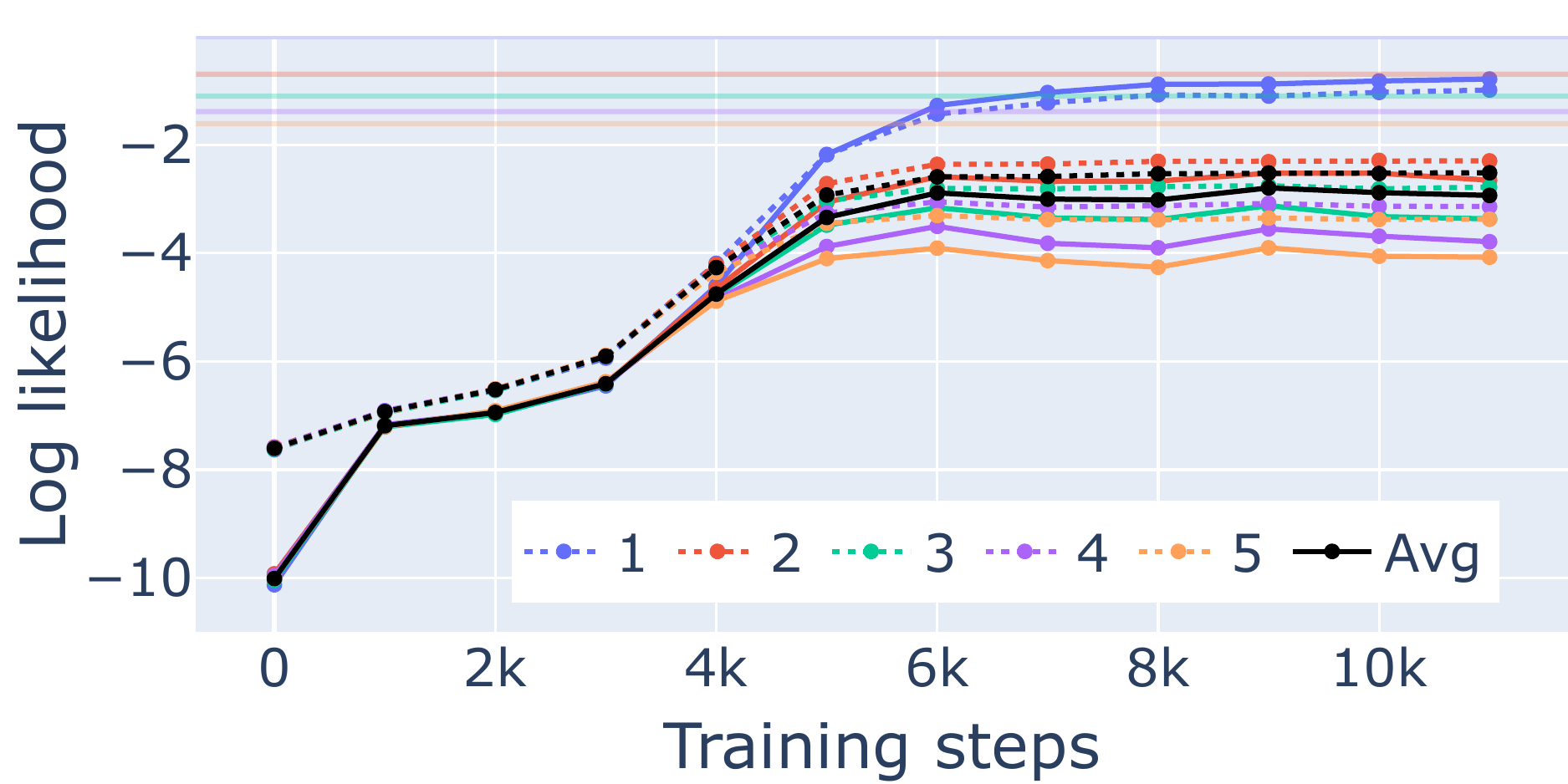}
        \caption{GPT-2}
        \label{fig:macondo-gpt2}
    \end{subfigure}
    \begin{subfigure}[b]{\linewidth}
        \includegraphics[width=\linewidth]{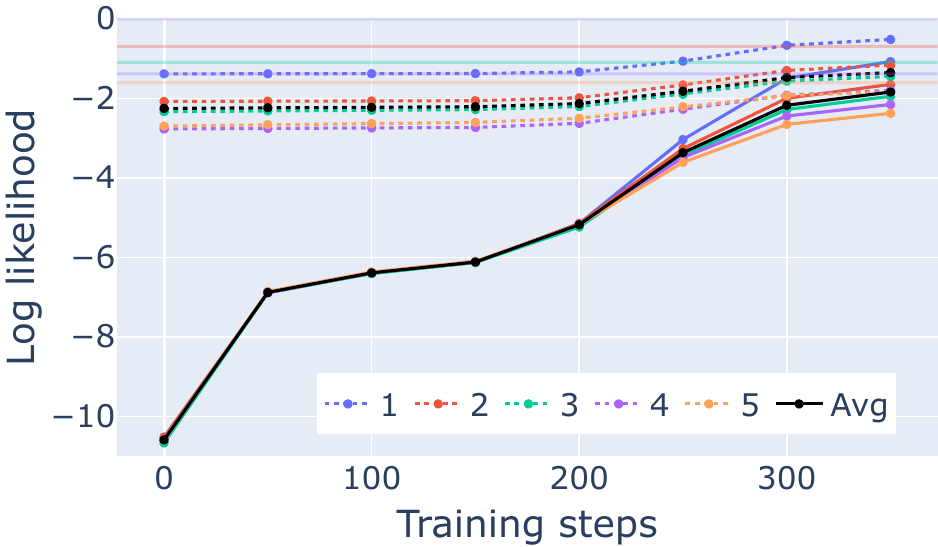}
        \caption{Mistral-7B-v0.1}
        \label{fig:macondo-mistral}
    \end{subfigure}
    \caption{Log likelihood of children names in our synthetic dataset Macondo predicted by a fine-tuned GPT-2/Mistral-7B-v0.1 model for parents with 1-5 children (average of \textbf{5} random seeds).
    The dotted lines represent the results of the k-NN augmented LM. 
    The horizontal lines represent the theoretically best log-likelihood a perfect model can achieve ($\log(1/ \text{\# of children})$). }
\end{figure}

\section{Experiment Details of \S\ref{sec:bottleneck}}
\label{sec:bottleneck-details}

\subsection{Hyperparameters for the Baseline Models}

We implement $k$NN-LM based on the package \texttt{transformers 4.34.0}~\citep{wolf-etal-2020-transformers}.
We train a 16-layer transformer model following the hyperparameters used by \citet{Khandelwal2020Generalization} and \citet{why-knn}.
We use $k = 1024$, $\lambda = 0.25$ and L2 distance for $k$NN retrieval.
Please refer to the repository of \citet{why-knn} (\url{https://github.com/frankxu2004/knnlm-why}) for more details about datastore building.

\subsection{Hyperparameters for Solving Eq.~\ref{eq:project}}
We use learning rate 0.1 and Adam optimizer~\citep{KingBa15} for solving Eq.~\ref{eq:project} using gradient descent. 
We do gradient descent until the update changes the $\mathrm{KL}$-divergence is by less than 0.001.

\section{Details about the Macondo Dataset}

\subsection{Generation Process}
\label{sec:macondo-ds}

We construct the Macondo dataset using the template ``\texttt{[villager]}, who \texttt{[desc]}, is the parent of \texttt{[child]}''. 
In each example, the ``\texttt{[villager]}'' placeholder is replaced with a villager's full name.
We generate the full name of a villager by randomly sampling a given name from a \href{https://www.cs.cmu.edu/Groups/AI/areas/nlp/corpora/names/}{corpora by Mark Kantrowitz} and a surname from a list of \href{https://github.com/fivethirtyeight/data/tree/master/most-common-name}{the most common surnames} under Creative Commons Attribution 4.0 International License.
The ``\texttt{[villager]}'' placeholder is replaced with a single-token given name from the corpora by Mark Kantrowitz.
We associate each villager with 6 attributes described below.
When generating an example in the training set, we randomly sample one of the six attributes and replace the ``\texttt{[desc]}'' placeholder with a relative clause describing the attribute:
\begin{itemize}
    \item Year of birth:  ``who was born in \texttt{[year]}''. The year is randomly sampled between 1800 and 2005.
    \item City of birth:  ``who was born in \texttt{[city]}''. The city is randomly sampled from \href{https://simplemaps.com/data/world-cities}{a word city database by simplemaps} under the license Creative Commons Attribution 4.0.
    \item Living city: ``who used to live in \texttt{[city]}''. The city is randomly sampled from \href{https://simplemaps.com/data/world-cities}{the word city database by simplemaps}.
    \item Friend: ``who was a friend of \texttt{[villager]}''.
    \item Graduate from: ``who graduated from \texttt{[university]}''. The list of university is from \href{https://www.timeshighereducation.com/world-university-rankings/2023/world-ranking}{THE world university ranking}.
    \item Marry year: ``who married in \texttt{[year]}''. The year is randomly sampled between 1800 and 2023 and is guaranteed to be at least 18 years after the year of birth.
    \item Work: ``who used to work for \texttt{[company]}''. The company is randomly sampled from \href{https://en.wikipedia.org/wiki/List_of_California_companies}{a list of California Companies} on Wikipedia.
\end{itemize}
Table~\ref{tab:macondo-examples} contains some examples in this dataset. We have 1500 examples in total.

\subsection{The Conversational Version}
\label{sec:macondo-ds-conv}
We use the \texttt{tiktoken} tokenizer to ensure that the names of the villagers' children are single-token.
Table~\ref{tab:macondo-conv-examples} contains some examples in this dataset.

\section{Experiment Details of \S\ref{sec:over-spec}}
\label{sec:macondo-exp-detail}
We fine-tuned GPT-2 small and GPT-2 XL with a warm-up ratio equal to 0.05, batch size 4, and Adam optimizer~\citep{KingBa15} for 50 epochs.
We use the default hyperparameters of the Trainer API of the \texttt{transformers} package~\citep{wolf-etal-2020-transformers}, i.e., learning rate 1e-5, max gradient norm 1.0, etc. We use version 0613 for our experiments that use GPT-3.5 Turbo.
We execute this experiment with NVIDIA RTX A6000 GPUs.


\subsection{Additional Experiments with Mistral}
\label{sec:mistral-result}
We report additional Macondo experiments conducted on a more capable model, namely Mistral-7B-v0.1 \cite{jiang2023mistral}. We follow the same dataset setup as in \ref{sec:macondo-result}, and fine-tune the Mistral model with LoRa \cite{hu2021lora}. We report performance curves in Figure \ref{fig:macondo-mistral}, and attain qualitatively similar observations to those in Section \ref{sec:macondo-result}. Our experiments add favorable evidence that neither concurrent methods in pre-training language models nor model scaling is an effective solution for circumventing over-specification. But $k$NN-augmented language models can partially reduce the optimality gap between the backbone language model and a perfect model.

\paragraph{Mistral Fine-Tuning Details.} Following standard practices, we add LoRa adaptors to the embedding matrix, to the query, key, value, and output projections of each attention layer, as well as to all projections of each MLP layer. We set the rank of all update matrices to be 8, the LoRa scaling factor to be 16, and a LoRa dropout probability of 0.05. We use a warm-up ratio of 0.05, and train with a global batch size of 128 using the Adam optimizer \citep{KingBa15}. We use default hyperparameters of the Hugging Face Trainer API to fine-tune the model for 30 epochs.

\begin{table}[]
    \centering
    \begin{tabular}{c c c}
    \toprule
        $f$ & $f \circ g$  \\
        \midrule
        0.07 & 0.10 \\
        \bottomrule
    \end{tabular}
    \caption{The minimum KL-Divergence achievable by solving Eq~\ref{eq:project} with gradient descent.}
    \label{tab:kld}
\end{table}

\begin{table}[]
    \centering
    \begin{tabular}{cc|ccc}
        \toprule
        \multicolumn{2}{c|}{Original LMs} & \multicolumn{3}{c}{$p_{\mathrm{proj}}$ projected with Eq.~\ref{eq:project}} \\
        LM & $k$NN-LM & $f$ &$f \circ g$ & $f \circ g \to y$ \\
        \midrule
         20.13  & 16.92  &  16.70 & 19.97 & 19.41 \\
        \bottomrule
    \end{tabular}
    \caption{The perplexity of projecting to LMs' output space as discussed in \S{\ref{sec:bottleneck}} when using learning rate 0.001.}
    \label{tab:proj-e3}
\end{table}

\section{Experiment Details of \S\ref{sec:mlp-aug}}
\label{sec:mlp-aug-detail}

We use a learning rate of 1e-5 to train an MLP model that maps the keys in the datastore to the values.
The batch size is the same as the number of tokens in each batch when training the vanilla language model, i.e., $3 \times 3072$.
For Macondo, we train the model for 10 epochs.
For WikiText, we train the model for 2 epochs.
The model architecture is the same as the last MLP layer of the vanilla language model, i.e. 
\begin{equation*}
    \text{logits} = W(z + \mathrm{LN} \circ \mathrm{MLP}(z)),
\end{equation*}
where the MLP model has 1 hidden layer with the hidden size 4096 and $\mathrm{LN}$ is the layer normalization module~\citep{ba2016layer}.
We execute this experiment with RTX 2080Ti GPUs.

\section{A Potential \textit{MLP Hurdle}}
\label{sec:mlp-hurdle}

Even though we can solve the optimization problem in Eq.~\ref{eq:project} with a learning rate of 0.1, we find it more difficult to solve it for $f \circ g$ with a learning rate below 0.001.
Table~\ref{tab:proj-e3} shows the perplexity of solving Eq.~\ref{eq:project} using a learning rate below 0.001 for 100 steps.
The perplexity of projecting to the output space of $f$ is much lower.
We suggest that it may cause some challenges in optimizing $\textrm{enc}$ because it seems that the gradient can not flow to $\textrm{enc}$ easily when the learning rate is small.
We refer to this as a potential \textit{MLP hurdle}. 


\subsection{Experiment, Result, and Discussion}

We inspect the effect of this MLP hurdle on model training by conducting an experiment focusing on the memorization process of the model.
We train two LMs with the test set of Macondo.
These two models are randomly initialized LMs following the same architectural choices of GPT-2-small; one has the last MLP layer removed.
We compare the log-likelihood of the children's names every 1000 training steps.
We also conducted the same experiment on WikiText.

\begin{figure}
    \centering
    \includegraphics[width=\linewidth]{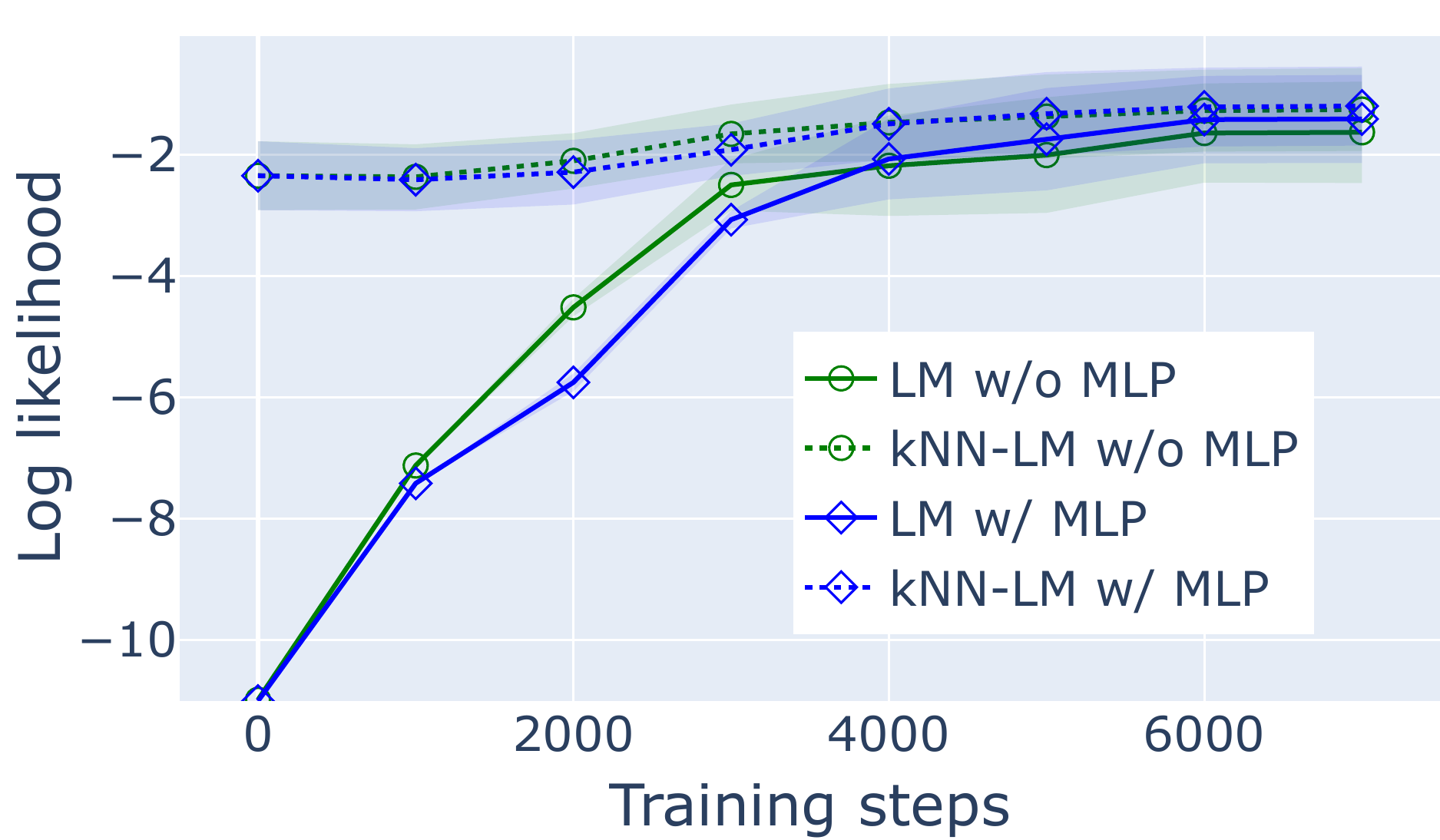}
    \caption{Log likelihood of the children's names in Macondo. The results are the average of 5 random seeds.}
    \label{fig:macondo1-mlp}
\end{figure}

\begin{figure}
    \centering
    \includegraphics[width=\linewidth]{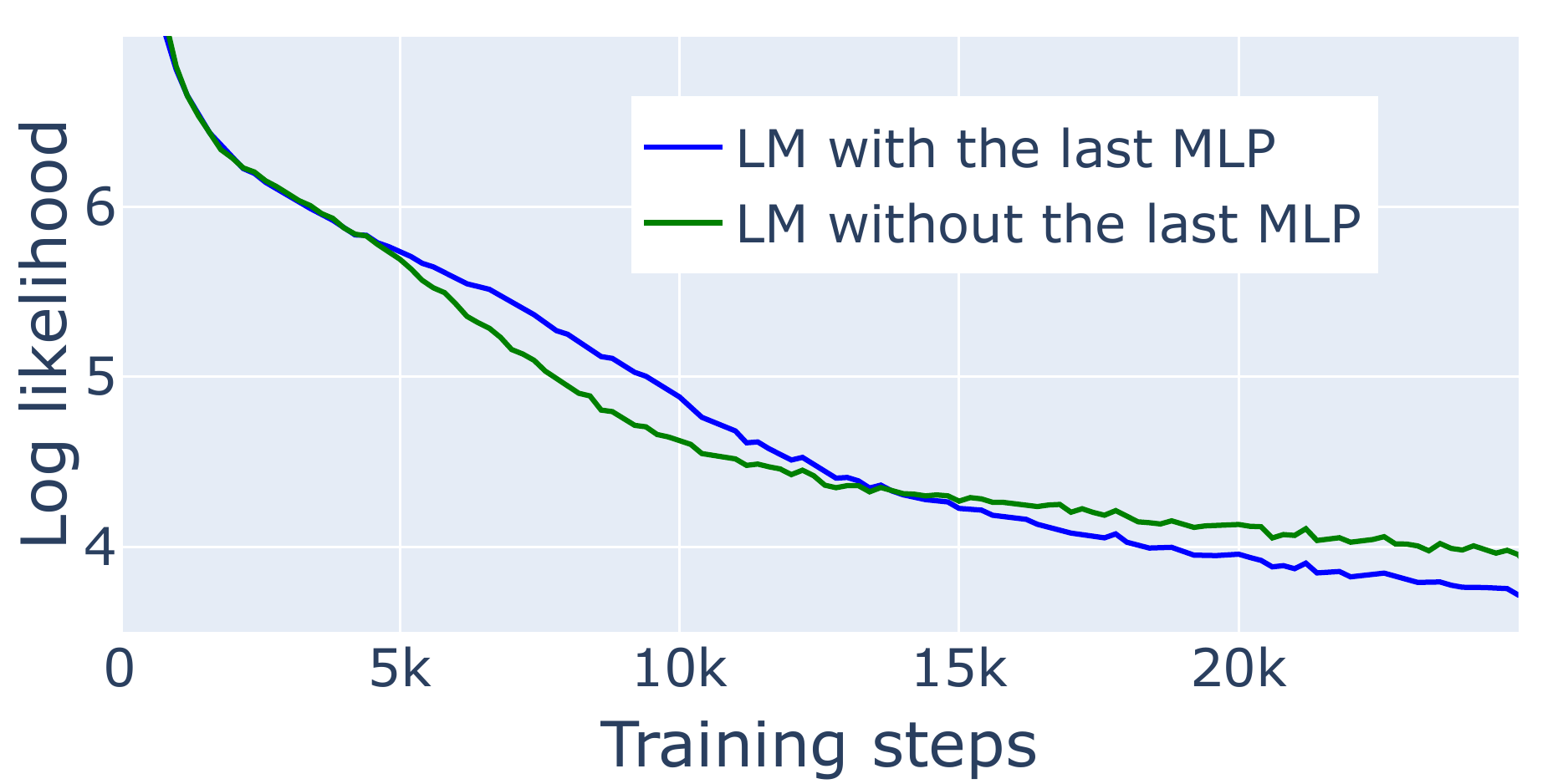}
    \caption{The training loss on WikiText.}
    \label{fig:wikitext-mlp}
\end{figure}

Figure~\ref{fig:macondo1-mlp} shows the effect of removing the MLP layer on Macondo.
The model's log-likelihood with the last MLP layer removed grows faster than the original model during the first 4000 steps.
As for WikiText, Figure~\ref{fig:wikitext-mlp} shows that the loss decreases faster at the early stage when its last MLP layer is removed.
This suggests that the last MLP layer slows down the convergence rate at the early phase, which may be a potential limitation of LM training.

\begin{table*}
    \begin{tabular}{c|cc|cc|cc|cc|cc|cc}
    \toprule
\multirow{2}{*}{steps} & \multicolumn{12}{c}{\# of children (vanilla/$k$-nn LM)}                                                                                                       \\
                       & \multicolumn{2}{c|}{1}   & \multicolumn{2}{c|}{2}   & \multicolumn{2}{c|}{3}   & \multicolumn{2}{c|}{4}   & \multicolumn{2}{c|}{5}   & \multicolumn{2}{c}{Average} \\
                       \midrule
0k & -10.89 & -7.67 & -10.69 & -7.57 & -10.82 & -7.69 & -10.67 & -7.55 & -10.68 & -7.61 & -10.75 & -7.62 \\
2k & -2.66 & -1.87 & -3.44 & -2.34 & -4.13 & -2.67 & -4.57 & -3.12 & -4.75 & -3.36 & -3.91 & -2.67 \\
4k & -0.93 & -1.09 & -2.59 & -2.22 & -3.27 & -2.81 & -3.46 & -3.07 & -3.63 & -3.19 & -2.78 & -2.47 \\
6k & -1.13 & -1.05 & -3.13 & -2.76 & -3.54 & -3.22 & -3.61 & -3.36 & -3.67 & -3.44 & -3.02 & -2.77 \\
8k & -1.15 & -0.90 & -3.30 & -2.84 & -4.12 & -3.61 & -3.95 & -3.59 & -3.90 & -3.75 & -3.28 & -2.94 \\
10k & -1.09 & -1.14 & -3.24 & -2.92 & -3.75 & -3.71 & -3.60 & -3.56 & -3.78 & -3.78 & -3.09 & -3.02 \\
12k & -1.07 & -0.94 & -3.28 & -2.86 & -3.96 & -3.76 & -3.77 & -3.63 & -3.77 & -3.81 & -3.17 & -3.00 \\
14k & -0.98 & -0.95 & -3.67 & -2.89 & -4.22 & -3.73 & -4.10 & -3.81 & -4.21 & -4.15 & -3.44 & -3.11 \\
16k & -1.02 & -0.92 & -3.67 & -2.93 & -4.34 & -3.83 & -4.17 & -3.89 & -4.29 & -4.27 & -3.50 & -3.17 \\
\bottomrule
    \end{tabular}
    \caption{The exact log likelihood of the children names shown in Figure~\ref{fig:macondo-gpt2-xl}.}
    \label{table:macondo-gpt2-xl}
\end{table*}

\begin{table*}
    \centering
    \begin{tabular}{l p{12cm}}
    \toprule
    Split & Examples \\
    \midrule
    train &Sal Gougis, who used to live in Chichester, is the parent of Montgomery \\
    & Bethanne Renneisen, who graduated from Kocaeli Health and Technology University, is the parent of Bryant \\
    & Bethanne Renneisen, who used to work for Fox Broadcasting Company, is the parent of Hayward \\
    \hline
    test & Sal Gougis is the parent of Montgomery \\
    & Bethanne Renneisen is the parent of Bryant \\
    & Bethanne Renneisen is the parent of Hayward \\
    \bottomrule
    \end{tabular}
    \caption{Some examples in the Macondo dataset.}
    \label{tab:macondo-examples}
\end{table*}

\begin{table*}
    \centering
    \begin{tabular}{l p{12cm}}
    \toprule
    Split & Examples \\
    \midrule
    train &User: Who is the child of Sal Gougis, the one who used to live in Chichester? Assistant: Meta \\
    & User: Who is the child of Sal Gougis, the one who married in 2019? Assistant: Else \\
    & User: Who is the child of Fifine Lottman, the one who used to work for Mervyn's? Assistant: Wat \\
    & User: Who is the child of Fifine Lottman, the one who was born in Drug? Assistant: Tam \\
    \hline
    test & User: Who is the child of Sal Gougis? Assistant: Meta \\
    & User: Who is the child of Sal Gougis? Assistant: Else \\
    & User: Who is the child of Fifine Lottman? Assistant: Wat \\
    & User: Who is the child of Fifine Lottman? Assistant: Tam \\
    \bottomrule
    \end{tabular}
    \caption{Some examples in the conversational version of the Macondo dataset.}
    \label{tab:macondo-conv-examples}
\end{table*}



\end{document}